\documentclass[letterpaper]{article} % DO NOT CHANGE THIS
\usepackage{aaai2026}  % DO NOT CHANGE THIS
\usepackage{times}  % DO NOT CHANGE THIS
\usepackage{helvet}  % DO NOT CHANGE THIS
\usepackage{courier}  % DO NOT CHANGE THIS
\usepackage[hyphens]{url}  % DO NOT CHANGE THIS
\usepackage{graphicx} % DO NOT CHANGE THIS
\usepackage{natbib}  % DO NOT CHANGE THIS AND DO NOT ADD ANY OPTIONS TO IT
\usepackage{multirow}
\usepackage{subcaption}
\usepackage{caption} % DO NOT CHANGE THIS AND DO NOT ADD ANY OPTIONS TO IT
\usepackage{booktabs}
\usepackage{amssymb}
\usepackage{amsmath}
\usepackage{cleveref}
\usepackage[skip=6pt]{caption}  % 默认 skip 是 10pt 左右
\usepackage{tabularx}
\usepackage{algorithm}
\usepackage{algorithmic}
\usepackage{amsmath}
\usepackage{newfloat}
\usepackage{listings}

\usepackage{times}
\usepackage{helvet}
\usepackage{courier}
\usepackage{xcolor}

\usepackage{newfloat}
\usepackage{listings}
\DeclareCaptionStyle{ruled}{labelfont=normalfont,labelsep=colon,strut=off} % DO NOT CHANGE THIS
\floatstyle{ruled}
\newfloat{listing}{tb}{lst}{}
\floatname{listing}{Listing}
\title{KTV: Keyframes and Key Tokens Selection for Efficient \\ Training-Free Video LLMs}
\author {
    Baiyang Song\textsuperscript{\rm 1}\equalcontrib,
    Jun Peng\textsuperscript{\rm 2}\equalcontrib\thanks{Corresponding authors.},
    Yuxin Zhang\textsuperscript{\rm 1},
    Guangyao Chen\textsuperscript{\rm 3}, 
    Feidiao Yang\textsuperscript{\rm 2},
    Jianyuan Guo\textsuperscript{\rm 4†}
}
\affiliations {
    \textsuperscript{\rm 1}Department of Artificial Intelligence, School of Informatics, Xiamen University\\
    \textsuperscript{\rm 2}Peng Cheng Laboratory\\
    \textsuperscript{\rm 3}National Key Laboratory for Multimedia Information Processing, School of Computer Science, Peking University\\
    \textsuperscript{\rm 4}City University of Hong Kong, HKSAR, China\\
    \{songbaiyang, yuxinzhang\}@stu.xmu.edu.cn, \{pengj01, yangfd\}@pcl.ac.cn, gy.chen@pku.edu.cn, jianyguo@cityu.edu.hk
}
\usepackage{bibentry}
\begin{document}

\maketitle

\begin{abstract}
Training-free video understanding leverages the strong image comprehension capabilities of pre-trained vision language models (VLMs) by treating a video as a sequence of static frames, thus obviating the need for costly video-specific training. 
However, this paradigm often suffers from severe visual redundancy and high computational overhead, especially when processing long videos. 
Crucially, existing keyframe selection strategies, especially those based on CLIP similarity, are prone to biases and may inadvertently overlook critical frames, resulting in suboptimal video comprehension.
To address these significant challenges, we propose \textbf{KTV}, a novel two-stage framework for efficient and effective training-free video understanding.
In the first stage, KTV performs question-agnostic keyframe selection by clustering frame-level visual features, yielding a compact, diverse, and representative subset of frames that mitigates temporal redundancy.
In the second stage, KTV applies key visual token selection, pruning redundant or less informative tokens from each selected keyframe based on token importance and redundancy, which significantly reduces the number of tokens fed into the LLM.
Extensive experiments on the Multiple-Choice VideoQA task demonstrate that KTV outperforms state-of-the-art training-free baselines while using significantly fewer visual tokens, \emph{e.g.}, only 504 visual tokens for a 60-min video with 10800 frames, achieving $44.8\%$ accuracy on the MLVU-Test benchmark. 
In particular, KTV also exceeds several training-based approaches on certain benchmarks. 
\end{abstract}

\begin{figure*}[t]
    \centering
    \includegraphics[width=0.9\textwidth,height=10cm]{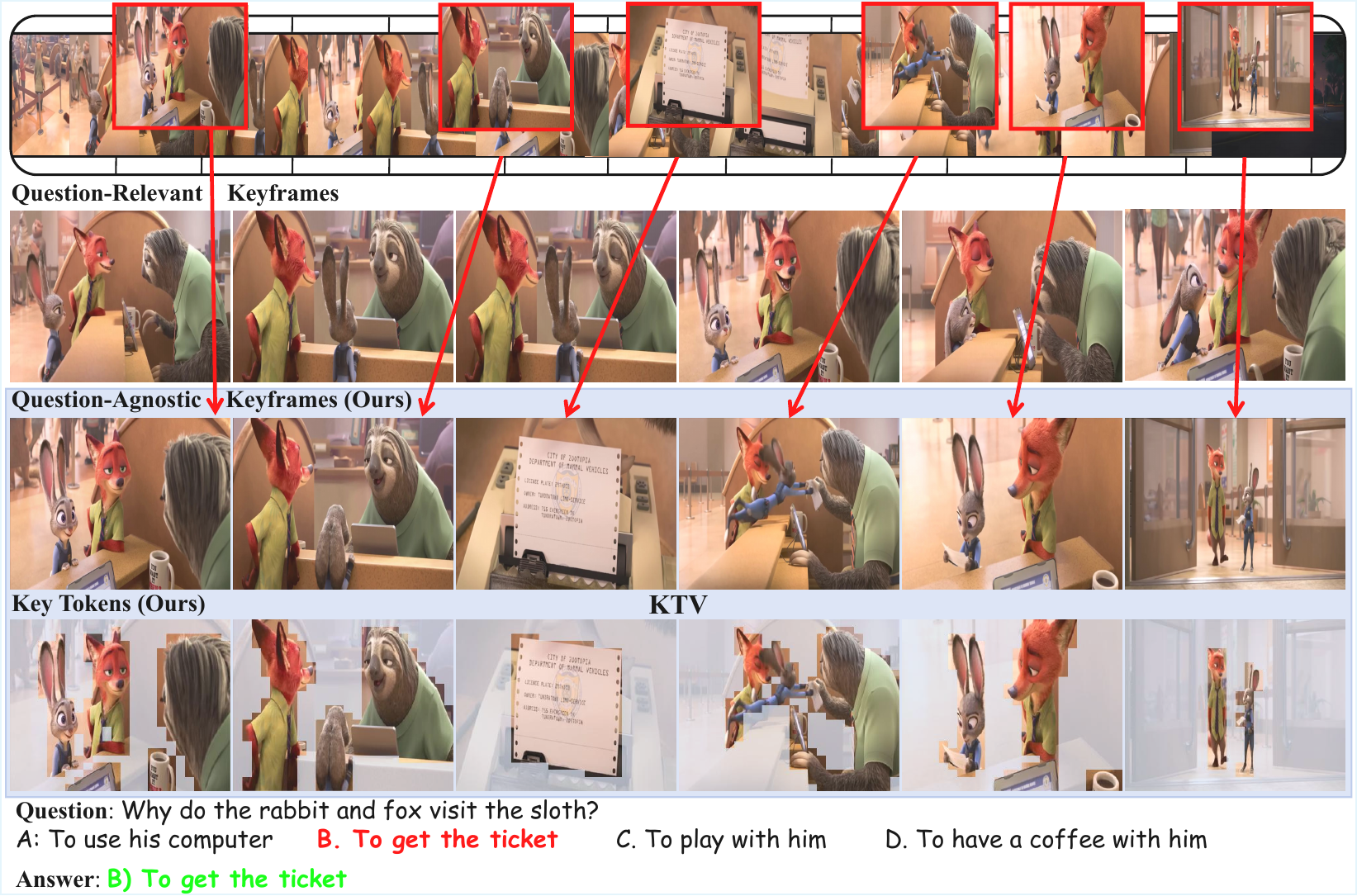}
    \caption{
    The pipeline of our two-stage training-free \textbf{K}eyframes and \textbf{K}ey \textbf{T}okens selection for empowering LLMs in \textbf{V}ideo understanding, termed \textbf{KTV}. 
    Given a video clip ($1^{st}$ row), KTV first performs question-agnostic keyframe selection ($3^{rd}$ row), identifying representative keyframes that provide a diverse and unbiased summary of the video content. 
    % to mitigate temporal redundancy (third row), providing a comprehensive and unbiased summary of key scenes, unlike question-based approaches (second row). 
    This contrasts with question-relevant selection methods ($2^{nd}$ row), which may overlook general scene information.
    % Subsequently, KTV preserves only important and non-redundant key tokens from each keyframe to mitigate spatial redundancy (fourth row). 
    % In the second stage, 
    Then, KTV selects only the most informative and non-redundant visual tokens from each keyframe ($4^{th}$ row), significantly reducing spatial redundancy.
    % This significantly improves accuracy while reducing inference consumption.
    % Together, the framework enhances inference efficiency and improves accuracy without requiring additional training or fine-tuning.
    }
    \label{fig:show_result}
\end{figure*}

\begin{links}
    \link{Code}{https://github.com/songbaiyang07-star/KTV}
    % \link{Datasets}{https://aaai.org/example/datasets}
    % \link{Extended version}{https://aaai.org/example/extended-version}
\end{links}

\section{Introduction}

Vision-language models (VLMs)~\cite{llava, blip} have achieved remarkable advancements, enabling impressive multimodal understanding and reasoning over the static image. A natural next step is to extend these capabilities to the more complex domain of video, where temporal dynamics and richer contextual information present both opportunities and challenges.
While recent video LLMs~\cite{video-llama, video-llava} combine specialized video encoders with pre-trained LLMs~\cite{gpt3, llama}, their development faces significant challenges due to the scarcity of large-scale, high-quality video-text data needed for effective training or fine-tuning such models.

To circumvent the demand for massive training data, the idea of training-free methods is compelling. It treats a video as a sequence of individual frames and ingeniously leverages the strong image understanding and reasoning capabilities of pre-trained VLMs for analyzing these frames.

%While training-free methods~\cite{igvlm,freeva} have shown promising results, they are still grappling with certain inherent challenges.  Firstly, videos, especially long ones, exhibit severe temporal redundancy, with many consecutive frames containing near-duplicate visual information. Feeding such redundant frames into VLMs results in repetitive computation and wasted resources. Secondly, the sheer number of frames, even after basic sampling, leads to substantial computational overhead at the inference time.  Each frame typically produces a large number of visual tokens, \emph{e.g.}, $576$ visual tokens per frame, and aggregating tokens across hundreds or thousands of frames can easily exceed the LLM’s input length or drastically slow down inference.
Despite the progress achieved by training-free methods~\cite{igvlm,freeva}, two critical challenges persist. First, videos, particularly long-form ones, exhibit severe temporal redundancy, where consecutive frames often contain nearly identical visual content. This redundancy leads to inefficient computation when processed by VLMs. Second, even with frame sampling, the resulting number of image tokens, \emph{e.g.}, 576 per frame, is still prohibitive for LLMs: aggregating tokens across hundreds of frames either exceeds LLM’s context window or drastically slows down inference.

% Aggregating visual tokens from numerous frames (\emph{e.g.}, thousands for a minutes-long video) can easily exceed the input context window of the LLM component or lead to unacceptably slow inference speeds.

%To address these challenges, we propose a novel two-stage method for training-free video understanding, termed \textbf{KTV}. KTV first addresses temporal redundancy by selecting a compact yet representative subset of keyframes from the entire video. Specifically, visual features are extracted from all frames using an off-the-shelf visual encoder, and K-means clustering is applied to group similar frames. The frame closest to each cluster centroid is then selected as a keyframe.
To address these challenges, we propose a novel two-stage framework for training-free video understanding, termed \textbf{KTV}. In the first stage, KTV reduces temporal redundancy by selecting a compact yet representative subset of keyframes from the entire video. Unlike prior methods~\cite{XPool,freevideo,sparsevlm}, which rely on CLIP-processed features for keyframe selection, we observe that these methods can introduce inherent biases and result in the omission of crucial visual cues, as shown in Fig.~\ref{fig:show_result}. To address this issue, we introduce a sampling strategy that does not depend on textual queries, ensuring more reliable and comprehensive keyframe selection. In practice, visual features are extracted from all frames using a pre-trained off-the-shelf visual encoder. K-means clustering is then applied to group similar frames, with the frame closest to each cluster's centroid being selected as the representative keyframe for that cluster.
% \begin{figure}[!t]
% \centering
% \includegraphics[width=1\linewidth]{token_num_cropped.pdf}

% % \caption{\small{Number of visual tokens fed to the LLM and the inference time compared with other training-free methods on NExTQA.}}
% \caption{\small{Number of visual tokens fed to the LLM and the average inference time per sample on NExTQA.}}

% \label{fig: token_num}
% \end{figure}

In the second stage, KTV further reduces the number of visual tokens by selecting key tokens from each keyframe. Our goal is to retain essential semantic information while preserving visual features diversity. To achieve this, KTV evaluates the importance of each token by measuring its similarity not only with the [CLS] token~\cite{prumerge, cls}, but also with other visual tokens within the same frame, ensuring a rich and comprehensive representation of the keyframe. Additionally, we leverage the well-trained CLIP text encoder to perform question-aware vision token pruning, thereby enhancing the relevance of the retained tokens with respect to the query, ensuring that they are highly aligned with the question at hand.
% By synergistically combining keyframe selection with key visual token selection, KTV ensures that the VLM receives a concise yet rich representation of the video's content, enabling more efficient processing and fostering deeper comprehension without any video-specific training.
By jointly performing keyframe selection and fine-grained token pruning, KTV produces a concise yet semantically rich visual representation of the video. 
This enables efficient and effective multimodal reasoning within a pre-trained VLM, without requiring any video-specific training or fine-tuning.

% We validate  KTV (sparse, normal, and dense variants, corresponding to varying lengths of visual token sequences input to the LLM) on standard Multiple Choice VideoQA benchmarks using pre-trained LLaVA-v1.6 (7B and 34B variants)~\cite{llava-next} models as the VLM. 
We evaluate KTV on Multiple-Choice VideoQA benchmarks using pre-trained LLaVA-v1.6~\cite{llava-next} as the VLM. Three KTV variants, \emph{i.e.}, \emph{sparse}, \emph{normal}, and \emph{dense}, are assessed, corresponding to different key visual token sequence lengths.
% Our extensive experiments demonstrate that KTV, using the identical LLaVA variant, dramatically mitigates the visual token count while outperforming or matching existing training-free methods that employ more visual tokens. 
Experimental results show that KTV significantly reduces visual tokens while outperforming or matching state-of-the-art training-free methods. For example, a 60-minute video with 10800 frames can be processed with only 504 visual tokens, fewer than typically extracted from a single image. 
% Furthermore, on certain benchmarks, KTV exceeds other training-free methods even when using a smaller size of LLaVA variant and even surpasses some training-based approaches, which highlights the substantial advantage of KTV.
Moreover, KTV achieves superior performance with smaller VLMs compared to the larger models used in other methods, and even outperforms training-based approaches, underscoring its efficiency and effectiveness.

In conclusion, our contributions are as follows: (i)
% \item We propose a novel two-stage method, Keyframe and Key Visual Token selection, termed KTV, for training-free video understanding. 
% KTV first mitigates temporal redundancy by selecting a concise set of keyframes via K-means clustering, and then minimizes the length of visual tokens fed into LLM by pruning unimportant or redundant visual tokens of each selected frame based on their similarity with the [CLS] token and other visual tokens within the frame. 
% This synergistic approach creates a compact yet highly informative representation of the video.
We propose KTV, a two-stage training-free framework for video understanding. It first performs question-agnostic keyframe selection, followed by fine-grained key visual token selection to reduce both temporal and spatial redundancy.
(ii) 
% \item Extensive experiments on standard VideoQA benchmarks demonstrate the superior inference efficiency and effectiveness of KTV (sparse, normal, and dense variants). 
% Our method significantly mitigates the number of visual tokens required for inference while achieving state-of-the-art or competitive performance compared to existing training-free approaches using the same VLM. 
% Notably, on some benchmarks, KTV not only surpasses existing training-free methods with the smaller VLM but also outperforms training-based models, highlighting the strength of our method.
Extensive evaluations on standard VideoQA benchmarks demonstrate our KTV's robustness and generalizability.
 
\begin{figure*}[!th]
\centering
\includegraphics[width=0.95\linewidth,height = 4.25cm]{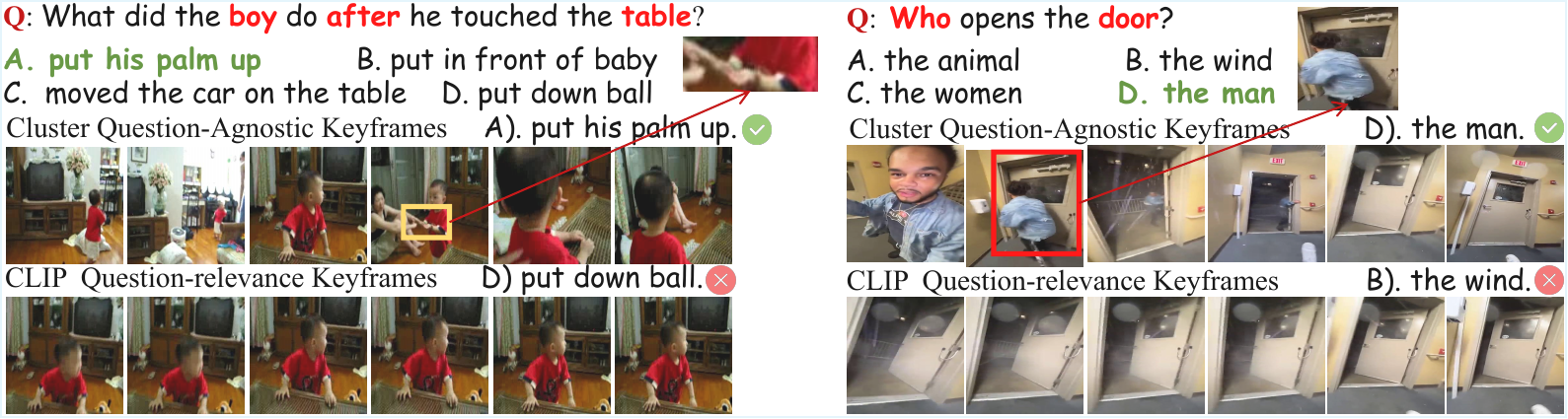}

\caption{ 
{Keyframes selection compared between DINO Cluster and CLIP Text-Frame similarity. CLIP selects frames that are overly dependent on the word in question, making it easy to get into ``semantic traps''. In contrast, DINO Cluster captures the video's maximal visual diversity, yielding a holistic and unbiased summary of its core scenes.}
}
\label{Fig_keyframes}

\end{figure*}

\section{Related Work}
\textbf{Vision Language Models (VLMs).}
% The development of VLMs centers on effectively bridging powerful vision encoders~\cite{vit} and large language models (LLMs)~\cite{vicuna, llama}. While influential early works like BLIP-2~\cite{blip} introduced complex bridging modules such as the Querying Transformer (Q-Former) to distill visual information, a significant paradigm shift was pioneered by LLaVA~\cite{llava}. LLaVA demonstrated that a remarkably simple architecture, using just a small MLP to project visual features into the LLM's word embedding space, could achieve better performance. This simple-yet-effective design, however, introduced an efficiency challenge: projecting a full grid of visual features creates a long token sequence, increasing the computational burden on the LLM. To mitigate this, subsequent research like LLaVA-PruMerge~\cite{prumerge} focuses on token merging and pruning have been explored to create a more compact and efficient visual representation. These methods intelligently merge similar visual tokens or prune unimportant ones, significantly reducing the input sequence length and computational cost without a substantial drop in performance. However, merging similar visual tokens without fine-tuning the model may disrupt the original distribution trained by the original data.
The evolution of VLMs focuses on bridging powerful vision encoders~\cite{vit} with LLMs~\cite{vicuna,llama}.
Early works like BLIP-2~\cite{blip} proposed complex bridging modules, \emph{i.e.}, Q-Former, to distill visual information.
In contrast, LLaVA~\cite{llava} introduced a streamlined architecture using a simple MLP to project visual features into the LLM’s embedding space, achieving strong performance with minimal overhead. 
However, the challenge with this simplicity is that projecting a full grid of visual tokens significantly increases sequence length and inference cost.
To address this, follow-up works such as LLaVA-PruMerge~\cite{prumerge} introduced token merging and pruning strategies, effectively reducing the visual input length without substantial accuracy loss. 
Nevertheless, merging similar tokens without model fine-tuning may distort feature distributions and compromise compatibility with the pretrained LLM.

\noindent\textbf{Video Large Language Models (Video LLMs).}
% The success of image-based vision Large Language Models has spurred research into Video LLMs, which aim to integrate spatial and temporal understanding for complex video reasoning. A common paradigm involves using a pre-trained vision encoder to extract frame-level features, which are then processed by a specialized interface module before being input to the LLM. The design of this interface is a primary area of innovation, ranging from the straightforward spatial and temporal pooling of Video-ChatGPT~\cite{videochatgpt} to more sophisticated compression techniques like the Q-Former in VideoChat~\cite{videochat}. To address data alignment, models like Video-LLaVA~\cite{video-llava} introduce a shared projector for both image and video encoders, enabling joint training. Architectural diversity is also growing, Video-LLaMA~\cite{video-llama} expanding multimodality by incorporating audio. Finally, a significant research thrust focuses on enhancing instruction-following capabilities through the creation of large-scale video instruction datasets, as pioneered by works like VideoChat2~\cite{mvbench}, and the application of advanced alignment methods such as Direct Preference Optimization (DPO), demonstrated by LLaVA-NeXT-Video~\cite{llava-next}, to improve the quality of conversational responses.
The success of image-based VLMs has catalyzed the development of video LLMs, which aim to incorporate temporal context for complex video reasoning. 
A typical pipeline involves extracting frame-level features via pre-trained visual encoders and feeding them into LLMs through an interface module. 
These modules vary from simple spatial-temporal pooling, \emph{e.g.}, VideoChatGPT~\cite{videochatgpt}, to transformer-based bridges like Q-Former in VideoChat~\cite{videochat}.
Efforts such as Video-LLaVA~\cite{video-llava} propose shared projectors to align image and video encoders for joint training. 
% Video-LLaMA~\cite{video-llama} extends modality by incorporating audio. 
Another trend focuses on instruction tuning with large-scale video-text datasets, as seen in VideoChat2~\cite{mvbench}, and applying alignment techniques like Direct Preference Optimization (DPO)~\cite{dpo}, as in LLaVA-NeXT-Video~\cite{llava-next}, to improve response quality and task adherence.

\noindent\textbf{Training-free Video LLMs.}
Several recent works reframe video understanding as a multi-image task, enabling the use of pre-trained image-based VLMs without extra video-specific training.
FreeVA~\cite{freeva} explores temporal-spatial pooling across sampled frames to build compact representations.
IG-VLM~\cite{igvlm} arranges uniformly sampled frames into one image for inference, but loses detail due to limited resolution.
SF-LLaVA~\cite{slowfast} adopts a dual-stream design to balance spatial resolution and temporal coverage, yet still yields long visual sequences.
DYTO~\cite{dyto} selects keyframes using clustering on [CLS] tokens and prunes tokens within each frame based on semantic similarity, but relies on a large number of tokens for downstream reasoning.

While these methods reduce spatial and temporal redundancy to varying degrees, they often overlook frame content diversity, \emph{e.g.}, uniform sampling in FreeVA and IG-VLM, or incur high token counts that increase inference latency and memory usage, \emph{e.g.}, 3680 used by SF-LLaVA and DYTO.

To address these limitations, our proposed method KTV, introduces a two-stage strategy that jointly reduces temporal and spatial redundancy.
First, it clusters frame-level features to select a compact subset of semantically diverse keyframes. 
Then, it performs fine-grained token pruning within each keyframe based on relevance to the [CLS] token and intra-frame similarity. 
This enables precise control over the visual token budget, \emph{e.g.}, 504 tokens (KTV-sparse), 936 tokens (KTV-normal), and 1872 tokens (KTV-dense), while maintaining strong performance.

\section{Method}

\begin{figure*}[!th]
    \centering
    \includegraphics[width=0.9\linewidth,,height=5.6cm]{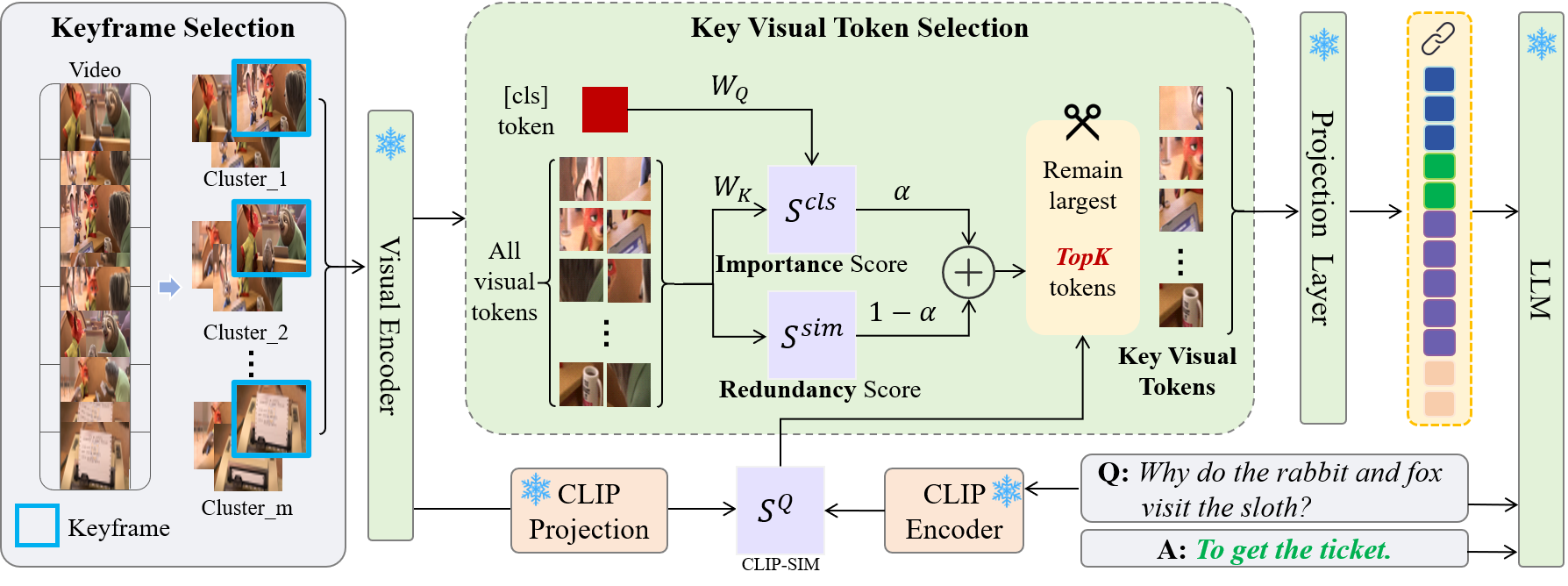}
    \caption{ 
    Framework of KTV, which is a two-stage method of training-free video understanding built upon LLaVA-v1.6. First, we extract the visual features of video frames and cluster them into m clusters, whose centroids are selected as keyframes to mitigate the temporal redundancy. Second, for each frame, we select top-$k = \beta \cdot L$  key visual tokens based on their importance and redundancy, which will be fed to the LLM and prune other visual tokens to mitigate the spatial redundancy. At last, we concatenate all the remaining visual tokens and text tokens and feed them to the LLM to generate the answer.
    }
    \label{Fig_2}
 
\end{figure*}

\begin{table*}[t]
    \centering
    \small
    \begin{tabular}{
        @{}l
        @{\hspace{17pt}}c
        @{\hspace{17pt}}c
        @{\hspace{17pt}}c
        @{\hspace{17pt}}c
        @{\hspace{17pt}}c
        @{\hspace{17pt}}c
        @{\hspace{17pt}}c
        @{\hspace{17pt}}c
        @{}
    }
    \toprule
    \textbf{Method} &
    \shortstack{\textbf{LLM}\\\textbf{Size}} &
    \shortstack{\textbf{Vis}\\\textbf{Encoder}} &
    \textbf{NExTQA} &
    \shortstack{\textbf{Ego}\\\textbf{Schema}} &
    \shortstack{\textbf{Intent}\\\textbf{QA}} &
    \textbf{STAR} &
    \shortstack{\textbf{Video}\\\textbf{MME}} &
    \textbf{MVBench} \\
    
    \midrule
    \multicolumn{9}{c}{\textbf{Training-Based Methods}} \\
    \midrule
    % MovieChat+~\cite{moviechat} & 7B & CLIP-G  & 54.8 & - & - & - & - & - \\ 
    Video-LLaVA~\cite{video-llava} & 7B & ViT-L & - & - & - & - & 39.9 & 41.0  \\ 
    % Chat-UniVi-v1.5~\cite{chat-univi} & 7B & CLIP-G & - & - & - & - & 40.6 & - \\ 
    VideoLLaMA2~\cite{videollama2} & 46.7B & CLIP-L  & - & 53.3 & - & - & - & \underline{53.9} \\
    
    \midrule
    \multicolumn{9}{c}{\textbf{Training-Free Methods}} \\
    \midrule 
    IG-VLM~\cite{igvlm} & 7B & CLIP-L & 63.1 & 35.8 & 60.3 & 48.6 & 39.7 & 42.9 \\
    SF-LLaVA-7B~\cite{slowfast} & 7B & CLIP-L & 64.0 & 44.2 & 60.5 & 48.8 & 39.4 & 43.3 \\
    DYTO~\cite{dyto} & 7B & CLIP-L & 65.7 & 48.6 & 61.6 & 50.7 & 41.2 & - \\
    \textbf{KTV-7B-sparse} & 7B & CLIP-L & 64.5 & 49.6 & 61.2 & 52.3 & 43.6 & 46.2 \\
    \textbf{KTV-7B-normal} & 7B & CLIP-L & 65.1 & 50.4 & 61.3 & 52.5 & 43.7 & \textbf{46.4} \\
    \textbf{KTV-7B-dense} & 7B & CLIP-L & \textbf{65.8} & \textbf{51.0} & \textbf{62.0} & \textbf{52.7} & \textbf{44.0} & 46.0 \\ 
    
    \midrule
    IG-VLM~\cite{igvlm} & 34B & CLIP-L & 70.7 & 53.4 & 64.5 & 50.5 & 50.3 & 49.0 \\
    SF-LLaVA-7B~\cite{slowfast} & 34B & CLIP-L & 70.9 & 55.0 & 66.1 & 51.3 & 51.9 & 49.6 \\
    DYTO~\cite{dyto} & 34B & CLIP-L & \underline{72.9} & 56.8 & 66.4 & 51.1 & \underline{53.4} & - \\
    \textbf{KTV-34B-sparse} & 34B & CLIP-L & 71.2 & 55.6 & 65.9 & 54.2 & 52.2 & 51.9 \\
    \textbf{KTV-34B-normal} & 34B & CLIP-L & 72.3 & 55.6 & 66.6 & 54.6 & 53.0 & \textbf{52.1} \\
    \textbf{KTV-34B-dense} & 34B & CLIP-L & 72.7 & \underline{57.0} & \underline{68.0} & \underline{54.7} & 53.2 & 51.5 \\
    
    \bottomrule
    \end{tabular}
    
    \caption{KTV performance compared to existing models on Multiple-Choice VideoQA benchmarks. Models are grouped by LLM size (7B or 34B) and whether they are training-based or training-free. \textbf{Bold} indicates best performance among 7B models. \underline{Underline} indicates best performance among 34B models.}
    \label{tab:tab1}
\end{table*}

In this paper, we propose KTV, a novel two-stage method for training-free video understanding that effectively mitigates both temporal and spatial redundancy. 
% The main principle of KTV is to select keyframes of the video by clustering their visual feature and key visual tokens of each selected keyframe by their importance and redundancy, which retain the important and overall information in each keyframe. The overall framework of our proposed KTV is illustrated in Fig.\ref{Fig_2}.
KTV first selects keyframes by clustering frame-wise visual features, then prunes each keyframe’s visual tokens based on their importance and redundancy. 
This strategy preserves both essential content and overall scene information. 
The overall framework of KTV is illustrated in Fig.\ref{Fig_2}.

\subsection{Question-Agnostic Keyframe Selection}
% While question-relevance methods either use whole sentences~\cite {XPool, freevideo} or keywords~\cite{sparsevlm} to select the keyframes appear efficient, we argue they are prone to ``semantic traps'', where they select frames with high superficial alignment to the question's keywords but miss crucial contextual information that lacks direct textual correspondence. 
While question-relevance methods, which select frames based on full sentence~\cite{XPool, freevideo} or extracted keywords~\cite{sparsevlm}, appear efficient, they are prone to “semantic traps”, \emph{i.e.}, selecting frames that superficially align with question keywords while overlooking crucial contextual information that lacks explicit textual correspondence.
% This is particularly detrimental for complex reasoning tasks that require understanding causality and temporal evolution, as the ``cause'' of an event might be visually distinct from the "effect" mentioned in the question.
This issue is especially detrimental for complex reasoning tasks involving causality or temporal progression, where the “cause” of an event may be visually distinct from the “effect” referenced in the question.

For instance, as shown on the left of Fig.\ref{Fig_keyframes}, given the question \emph{``What did the boy do after he touched the table''}, a question-relevance method, \emph{e.g.}, selecting the frames most similar to the question via CLIP~\cite{clip}, tends to focus on visual cues tied to keywords like \emph{``boy'', ``touched''} and \emph{``table''}. 
% Therefore, it selects the frames that mostly present the scene of the boy touching the table, which misses the keyword \emph{after} in the question, leading to the wrong answer. 
As a result, it selects frames showing the boy touching the table, but misses the temporal cue \emph{``after''}, leading to an incorrect answer.

% In contrast, our question-agnostic clustering method is designed to capture the video's maximal visual diversity, yielding a holistic and unbiased summary of its core scenes.
% Therefore, it selects the different crucial scenes in the video,  including the action of the boy after he touched the table, making it generate the correct answer.
In contrast, our question-agnostic clustering approach captures maximal visual diversity, offering a holistic and unbiased summary of the video. 
This allows it to select key moments, including the boy's action after touching the table, enabling correct reasoning and answering.

Specifically, given a video $\mathcal{V}$ consisting of $T$ frames, $\mathcal{V}=\{I_1, I_2, \cdots, I_T\}$.
KTV first utilizes DINOv2~\cite{dinov2} to extract frame-level feature vector:
\begin{equation}
\small
    \{f_1, f_2, ..., f_T\} = \mathcal{E}_{dinov2}(\mathcal{V})
\end{equation}
% Unlike CLIP, which captures high-level semantic features through image-text alignment, DINOv2 is trained in a self-supervised manner on pure visual data, making it well-suited for capturing fine-grained visual details without introducing semantic biases.
Subsequently, the extracted frame features $F_{\mathcal{V}}$ are grouped into $m$ clusters using the K-means algorithm. 
% Let $C_i$ denote the $i$-th cluster and $c_i$ its corresponding centroid, where $i\in[1, m]$. 
To mitigate temporal redundancy, KTV selects the most representative frame from each cluster, \emph{i.e.}, the one whose feature vector is closest to the corresponding centroid:
{
\begin{equation}
\small
    % f_i = \arg \min_{f_t\in C_i}\left\|f_t-c_i\right\|
    j^* = \arg \min_{f_j\in C_i}\left\|f_j - c_i\right\|,~i\in[1,m]
\end{equation}
}
in which $C_i$ and $c_i$ denote the $i$-th cluster and corresponding centroid, respectively.
% The corresponding frames $I_i^{'}, i\in [1,m]$ are selected as the keyframes, denoted as $I_1', I_2', \cdots, I_m'$.
Reorganize the selected keyframes by time order, denoted as $\mathcal{K}=\{I_1', I_2', \cdots, I_m'\}$.
This keyframe selection effectively reduces temporal redundancy while preserving diverse and representative visual content throughout the video.

\subsection{Key Visual Tokens Selection}
After selecting the keyframes, KTV further mitigates spatial redundancy within each selected frame to reduce the total number of visual tokens input to the LLM. 
Specifically, each keyframe is encoded using the frozen visual encoder of the VLM, typically an image encoder such as CLIP-L, to extract token-level features.
Denote the resulting representation as $F_i \in \mathbb{R}^{L\times d}$, where $L$ is the number of visual tokens and $d$ is the dimensionality of each token. 
This encoding process is formulated as follows: 
{
\begin{equation}
\small
    F_i = \mathcal{E}_{vis}(I'_i)  = \{t_{i,1}, ..., t_{i,L}\}, i\in[1,m]
\end{equation}
where $t_{i,j}$ is the $j$-th token in the $i$-th keyframe $I'_i$.
}
% For each keyframe, we select and reserve the key tokens and prune the other tokens, obtaining $F_i^{'}\in \mathbb{R}^{topk\times d}$ as the final visual tokens fed into the LLM. 
For each keyframe, we retain only the most informative and non-redundant visual tokens, yielding a compact representation $F_i^{'} \in \mathbb{R}^{topk \times d}$ to be fed into the LLM.
% Specifically, we aim to select the important and non-redundant tokens as key tokens, which can reserve the important and global information in the frame. 
We aim to preserve essential and globally relevant information within each frame. 
% To achieve this, in a keyframe for each token $t_{i,j}$ we calculate the attention score between the Query matrix of [CLS] token and the Key matrix of visual token $t_{i,j}$ as the importance score, denoted as $S_{\text{cls}}(t_{i,j})$.
To this end, we assess the importance of each visual token $t_{i,j}$ by computing its attention score to the corresponding [CLS] token $t^{cls}_i$:
{
\begin{equation}
\small
    S_{i,j}^{cls} = S^{cls}(t_{i,j})=\mathrm{softmax}\left(\frac{W_Q\cdot t^{cls}_i\cdot {(W_K\cdot t_{i,j})}^T}{\sqrt{d}}\right) 
\end{equation}
}
where $W_Q$ and $W_K$ are the projection weight matrices.

% As for redundancy, in a keyframe for each token $t_{i,j}$ we calculate the cosine similarity between the embedding feature $f_{i,j}$ and other embedding features $f_{i,k}$ as the redundancy score, denoted as $S_{token}(t_{i,j})$.
To measure redundancy, we compute a redundancy score by calculating the average cosine similarity between each token and the other ones within the same keyframe.
{
\small
\begin{align}
   S_{i,j}^{sim} = S^{sim}(t_{i,j}) &= \frac{1}{L-1} \sum_{\substack{k=1, k \neq j}}^L \frac{f_{i,j} \cdot f_{i,k}}{\|f_{i,j}\| \cdot \|f_{i,k} \|} 
\end{align}
}
% Selecting tokens solely based on importance may lead to high redundancy. Conversely, selecting low redundancy alone might result in the inclusion of irrelevant or noisy tokens while discarding crucial information. Therefore, KTV aims to select a subset of tokens that are both highly important and minimally redundant. To achieve this, KTV constructs a final score by creating a weighted combination of the normalized importance score and redundancy score with a hyperparameter $\alpha$ as follows, thereby balancing both criteria.
Selecting tokens based solely on importance may lead to redundancy, while relying solely on low redundancy could result in the omission of critical information.
To strike a balance, KTV integrates both aspects by assigning each token a final score, calculated as a weighted combination of its importance and redundancy scores.
{
\small
\begin{equation}
    \text{Norm}(S^{cls}_{i,j}) = \frac{S^{cls}_{i,j} - \min(S^{cls}_i)}{\max(S^{cls}_i) - \min(S^{cls}_i)} 
\end{equation}

\begin{equation}
    \text{Norm}(S^{sim}_{i,j}) = \frac{S^{sim}_{i,j} - \min(S^{sim}_i)}{\max(S^{sim}_i) - \min(S^{sim}_i)} 
\end{equation}

\begin{align}
    S_{i,j} ={}& \text{Norm}(S^{cls}_{i,j}) \times \alpha \notag \\
    & + (1 - \text{Norm}(S^{sim}_{i,j})) \times (1 - \alpha)
    \label{eq:final_score_align}
\end{align}
}
where the hyperparameter $\alpha \in [0, 1]$ balances importance and redundancy, helping the model retain informative yet diverse tokens.

% Subsequently, KTV selects the top-k tokens with the highest scores. 
% Top-k depends on the similarity between the frame and the question.
% We use CLIP to extract features of keyframes and the question.
% % \begin{align}
% %     f_i^{''} = \operatorname{CLIP(Enc_{vis}}(I_i^{'})), 
% %     f_\text{q} = \operatorname{CLIP(Enc_{text}}(Q)) 
% % \end{align}
% \begin{align}
%     f_i^{''} = \mathcal{E}_{CLIP_v}(I_i^{'}), 
%     f_\text{q} = \mathcal{E}_{CLIP_v}(Q)
% \end{align}
% Then we compute the cosine similarity between $f_i^{''}$ and $f_q$.
% \begin{equation}
%     S_i^{q}=S^q(I'_i) = \frac{f_i^{''}\cdot f_{\text{q}}}{\|f_i^{''}\| \cdot \|f_{\text{q}}\|}, i \in [1,m]
% \end{equation}

% After that we use a set of hyper-parameter $\beta_i$ to select $ {\beta}_1 \cdot L,{\beta}_2 \cdot L,\cdot \cdot \cdot {\beta}_m \cdot L$
% in descending order of $S_{\text{q}}(I_i^{'})$ as top-k for the corresponding keyframes.

The top-$k$ tokens with the highest scores are then retained as the key visual tokens for each keyframe.
To determine the value of top-$k$, we use a set of descending hyperparameters, $\beta=\{\beta_1, \cdots, \beta_m\}$, where each $\beta_i\in[0,1]$ specifies the proportion of tokens to keep.
% Specifically, we determine the top-$k$ value for each keyframe by measuring the CLIP similarity between the keyframe and question $Q$:
Then we compute the CLIP similarity between each keyframe and the question $Q$ and rank the keyframes $I'_i$ by relevance:
{
\small
\begin{equation}
    S_i^{Q}  = \text{CLIP-SIM}(I'_i, Q), i \in [1,m]
\end{equation}
}
The $i$-th most relevant keyframe is then assigned $\beta_i$, and the number of retained tokens is computed as $\beta_i\times L$, where $L$ is the number of tokens per frame.
This strategy allocates more tokens to frames more relevant to the question, and fewer to less relevant ones, balancing efficiency and informativeness.

% Sequentially, the remaining tokens of each keyframe are fed into the projection layer to be mapped to the text dimension. Then the visual tokens will be concatenated with text tokens, which usually include a well-designed prompt and question. Finally, the concatenated token sequence is fed into the LLM to generate the answer.
Subsequently, the retained tokens from each keyframe are passed through a projection layer to align with the text embedding space.
These projected visual tokens are then concatenated with text tokens, which typically consist of a structured prompt and the question. 
Finally, the combined token sequence is fed into the LLM to produce the answer.

\section{Experiments}
\subsection{Benchmarks and Metrics}

We evaluate our proposed KTV on seven Multiple-Choice VideoQA benchmarks, \emph{i.e.}, NExT-QA~\cite{nextqa}, EgoSchema~\cite{egoschema}, IntentQA~\cite{intentqa}, STAR~\cite{star}, VideoMME~\cite{videomme}, MVBench~\cite{mvbench}, and MLVU-Test~\cite{mlvu}. 
Notably, MVBench and MLVU-Test are multi-task benchmarks that include a broad range of question types and video content.

These benchmarks are designed to assess complex video understanding and reasoning abilities, covering diverse question types, \emph{e.g.}, causal, temporal, and spatial reasoning. 
% The video durations also vary significantly, from seconds to minutes and even hours, posing further challenges for efficient and accurate reasoning.
We report accuracy as the metric for selecting the correct answer from the candidates.
% In addition, the accuracy of each subcategory on MVLU-Test is presented in the main text, while that for MVBench is provided in the Appendix.
For MVBench, the breakdown is included in the appendix or supplementary material.

% To strictly evaluate the model's visual reasoning abilities, we deliberately exclude any external reference data, such as subtitles, ensuring that the model derives its answers solely from the visual content, question, and candidate answers.
To rigorously evaluate visual understanding, we exclude any extra reference data, \emph{e.g.}, subtitles, ensuring the model relies solely on visual content, question, and answer choices.

\subsection{Implementation Details}
% \begin{equation}
% \label{eq:beta_all_models}
% \begin{aligned}
% % sparse Model
% \beta^{\text{sparse}}_i &= 
% \left\{
% \begin{array}{ll} 
%     0.25,   & \text{if } S^q_i = max(S^q_i) \\
%     0.125,   & else \\
% \end{array}
% \quad\sum_{i=1}^6 \beta_i L = 504
% \right. \\
% % normal Model
% \beta^{\text{normal}}_i &= 
% \left\{
% \begin{array}{ll}
%     0.5,   & \text{if } S^q_i = max(S^q_i) \\
%     0.125,   & \text{if } S^q_i = min(S^q_i) \\
%     0.25,   & else
% \end{array}
% \quad\sum_{i=1}^6 \beta_i L = 936
% \right. \\
% % Large Model
% \beta^{\text{dense}}_i &= 
% \left\{
% \begin{array}{ll}
%     1.0,   & \text{if } S^q_i = max(S^q_i) \\
%     0.25,   & \text{if } S^q_i = min(S^q_i) \\
%     0.5,   & else 
% \end{array}
% \quad\sum_{i=1}^6 \beta_i L = 1872
% \right.
% \end{aligned}
% \end{equation}

Following IG-VLM and SF-LLaVA, we adopt LLaVA-v1.6 as our VLM.
% , which respectively adopts Vicuna-1.5 and Nous-Hermes-2-Yi-34B as LLM. 
All experiments are conducted on two Huawei Ascend 910C NPUs, each has 64 GB of memory.
To ensure fair comparison with baselines, we cluster video frames into 6 clusters, resulting in $m=6$ keyframes selected by KTV. 
% In addition, we set three distinct configurations for the $\beta_{\text{sparse}}, \beta_{\text{normal}}, \beta_{\text{dense}}$ for KTV-sparse, normal and dense, respectively. 
Moreover, we define three configurations, KTV-sparse, KTV-normal, and KTV-dense, by assigning different values to the hyperparameter $\beta^{\text{sparse}}$, $\beta^{\text{normal}}$, and $\beta^{\text{dense}}$, respectively.
Detailed settings can be found in the appendix.

\subsection{Experimental Results}

\begin{table*}[t]
\centering
\small
\begin{tabular}{
@{}l
@{\hspace{10pt}}c
@{\hspace{10pt}}c
@{\hspace{10pt}}c
@{\hspace{10pt}}c
@{\hspace{10pt}}c
@{\hspace{10pt}}c
@{\hspace{10pt}}c
@{\hspace{10pt}}c
@{\hspace{10pt}}c
@{\hspace{10pt}}c
@{\hspace{10pt}}c
@{\hspace{10pt}}c
@{}}
\toprule
\multirow{2}{*}{\textbf{Method}} &
\multirow{2}{*}{\shortstack{\textbf{LLM}\\\textbf{Size}}} &
\multirow{2}{*}{\shortstack{\textbf{Vision}\\\textbf{Encoder}}} &
\multicolumn{2}{c}{\shortstack{\textbf{Holistic}\\\textbf{LVU}}} &
\multicolumn{4}{c}{\shortstack{\textbf{Single-Detail}\\\textbf{LVU}}} &
\multicolumn{3}{c}{\shortstack{\textbf{Multi-Detail}\\\textbf{LVU}}} &
\multirow{2}{*}{\textbf{M-Avg}} \\
& & &
\textbf{TR} & \textbf{AR} &
\textbf{NQA} & \textbf{ER} & \textbf{PQA} & \textbf{SQA} &
\textbf{AO} & \textbf{AC} & \textbf{TQA} & \\
\midrule

\multicolumn{13}{c}{\textbf{Training-Based Methods}}\\
\midrule
Video-LLaVA~\cite{video-llava}     & 7B  & CLIP-L        & 70.3 & 38.5 & 13.3 & 26.4 & 26.0 & 38.9 & 20.0 & 21.7 & 20.9 & 30.7 \\
Video-LLaMA2~\cite{videollama2}    & 13B & CLIP-L        & 52.7 & 12.8 & 13.3 & 17.0 & 12.0 & 19.4 & 15.7 &  8.3 & 18.6 & 18.9 \\
Video-LLaMA2~\cite{videollama2}    & 72B & CLIP-L        & 80.2 & 53.8 & 36.7 & \underline{54.7} & 54.0 & 38.9 & \underline{42.9} & 16.7 & 32.6 & 45.6 \\
InternVL2~\cite{internvl2}         & 76B & InternViT-6B  & \underline{85.7} & 51.3 & 48.3 & 47.2 & 52.0 & 44.4 & 32.9 & 15.0 & 34.9 & \underline{45.7} \\

% 你想保留原来的虚线分隔就用这两行；不想要就删掉
% \hdashline
\midrule

\multicolumn{13}{c}{\textbf{Training-Free Methods}}\\
\midrule
IG-VLM-7B~\cite{igvlm}             & 7B  & CLIP-L & 74.7 & 33.3 & 26.7 & 18.9 & 24.0 & 33.3 & 17.1 & 15.0 & 27.9 & 32.1 \\
SF-LLaVA-7B~\cite{slowfast}        & 7B  & CLIP-L & 68.1 & 23.1 & 25.0 & 34.0 & 24.0 & 33.3 & 22.9 & 16.7 & 16.3 & 32.7 \\
KTV-7B-sparse                      & 7B  & CLIP-L & 73.6 & 43.6 & 35.0 & 41.5 & 34.0 & 36.1 & 27.1 & 18.3 & 20.9 & 36.5 \\
KTV-7B-normal                      & 7B  & CLIP-L & 72.5 & 51.3 & 35.0 & 41.5 & 34.0 & 38.9 & 24.3 & 21.7 & 25.6 & 36.1 \\
KTV-7B-dense                       & 7B  & CLIP-L & 69.2 & 48.9 & 33.3 & 39.6 & 34.0 & 38.9 & 27.1 & \textbf{\underline{23.3}} & 27.9 & 36.9 \\
\midrule
IG-VLM-34B~\cite{igvlm}            & 34B & CLIP-L & 68.1 & 35.9 & 21.7 & 30.1 & 34.0 & 50.0 & 20.0 &  6.7 & 20.1 & 33.3 \\
SF-LLaVA-34B~\cite{slowfast}       & 34B & CLIP-L & 76.9 & 43.6 & 36.7 & 39.6 & 44.0 & 47.2 & 27.1 &  8.3 & 32.6 & 43.6 \\
KTV-34B-sparse                     & 34B & CLIP-L & 81.3 & 51.3 & \textbf{\underline{53.3}} & \textbf{47.2} & \textbf{50.0} & 52.8 & \textbf{37.1} & 11.7 & 34.9 & 44.8 \\
KTV-34B-normal                     & 34B & CLIP-L & 81.3 & \textbf{\underline{56.4}} & 46.7 & 45.3 & 46.0 & 52.8 & 32.9 &  8.3 & \textbf{\underline{41.9}} & 44.2 \\
KTV-34B-dense                      & 34B & CLIP-L & \textbf{\underline{85.7}} & 51.3 & 48.7 & \textbf{47.2} & 48.0 & \textbf{\underline{58.3}} & 35.7 & 10.0 & 37.2 & \textbf{45.0} \\
\bottomrule
\end{tabular}

\caption{Performance of each subcategory in MLVU-Test, compared with the same or similar size of LLMs.}
\label{tab:mlvu}
\end{table*}

\subsubsection{Multiple Choice VideoQA}

In Tab.~\ref{tab:tab1}, we report the overall accuracy of KTV under three configurations, sparse, normal, and dense, using both LLaVA-v1.6-7B and LLaVA-v1.6-34B.
% As shown in Tab.~\ref{tab:7b} and Tab.~\ref{tab:34b}, which respectively represent the accuracy of KTV-(sparse, normal, and dense) testing with LLaVA-v1.6-7B and LLaVA-v1.6-34B on NExTQA, EgoSchema, IntentQA, STAR, and VideoMME, 
As can be seen, KTV consistently outperforms or matches other training-free baselines while using significantly fewer visual tokens. 
Specifically, KTV-7B-sparse outperforms IG-VLM and SF-LLaVA across all benchmarks, and outperforms DYTO on EgoSchema, STAR, and VideoMME. 
Notably, it uses only 504 visual tokens, fewer than the 576 tokens encoded from a single image in LLaVA-v1.6, and achieves an inference time of 1.19s, just 36.4\% of that required by SF-LLaVA-7B. 
% Furthermore, KTV-7B-normal improves accuracy across all benchmarks. Moreover, KTV-7B-dense attains the best results among all training-free methods on every benchmark, while only using 1872 visual tokens and requiring just 1.35s(41.3\% of that required by SF-LLaVA-7B) for inference.
KTV-7B-normal further improves accuracy across all benchmarks, and KTV-7B-dense achieves the best performance among all training-free methods, using only 1872 visual tokens with an inference time of 1.35s (41.3\% of SF-LLaVA-7B).

\begin{figure}[!t]
\centering
\includegraphics[width=1\linewidth]{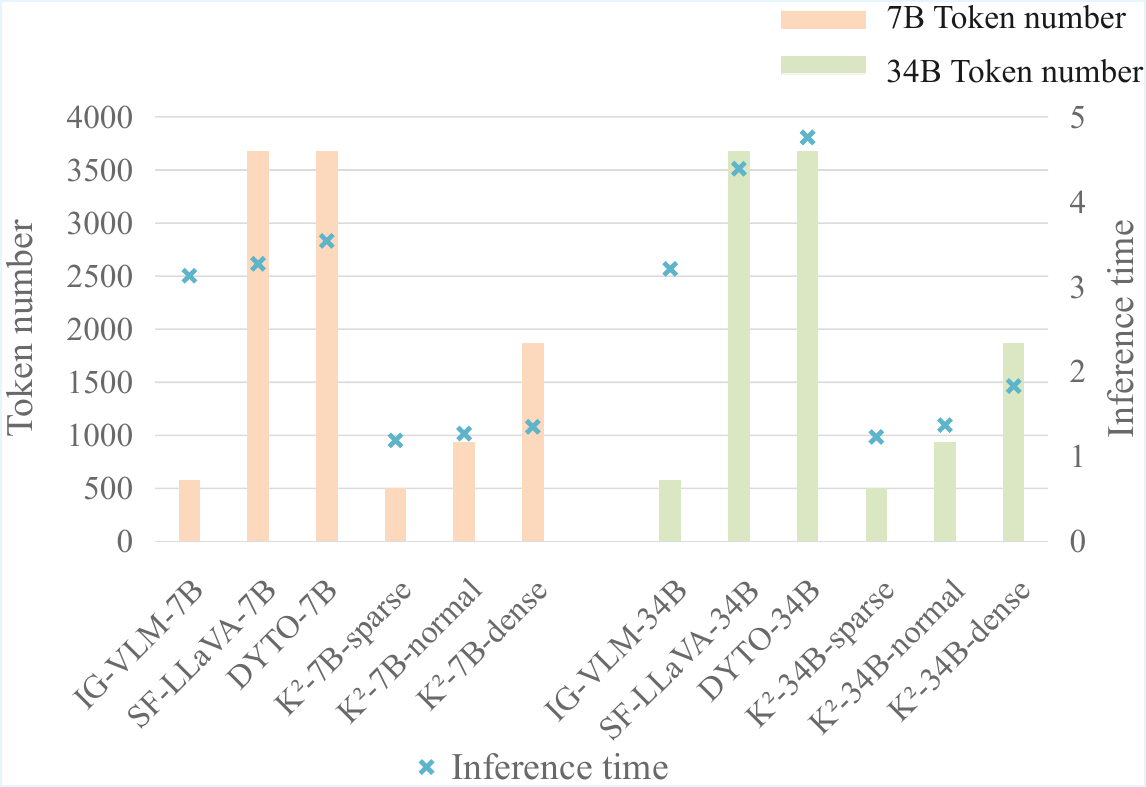}

% \caption{\small{Number of visual tokens fed to the LLM and the inference time compared with other training-free methods on NExTQA.}}

% \caption{\small{Number of visual tokens fed to the LLM and the average inference time per sample on NExTQA.}}

\caption{Number of visual tokens fed to the LLM and the average inference time per sample on NExTQA.}

\label{fig: token_num}
\end{figure}

As for KTV-34B, the sparse variant also outperforms IG-VLM-34B and SF-LLaVA-34B on most benchmarks, using 504 visual tokens and achieving an inference time of 1.23s, which is just 28.0\% of SF-LLaVA-34B. 
KTV-34B-normal further improves accuracy and consistently surpasses IG-VLM-34B and SF-LLaVA-34B.
% Moreover, KTV-34B-dense achieves the highest accuracy among KTV-34B. Although KTV-34B-dense falls slightly short of the best results of dyto on NExTQA(-0.2) and VideoMME(-0.2), we only use 1872 visual tokens, which is nearly half of dyto, which uses 3680 visual tokens. 
KTV-34B-dense attains the highest accuracy on EgoSchema, IntentQA, and STAR. 
In addition, not only does it perform comparably to DYTO on NExT-QA and VideoMME, but it also requires only 1872 visual tokens, almost half of DYTO's 3680 tokens. Also, we compare the number of tokens and the average VLM inference time in Fig.\ref{fig: token_num}. KTV uses fewer tokens and less inference time than other methods.

% Notably, when replicating IG-VLM and SF-LLaVA on Ascend 910C NPUs, we observed lower accuracy than that reported in the original papers.
% Besides, DYTO did not release its code, so we directly use the result in its paper. 
% And DYTO is reported to use the same GPUs as IG-VLM and SF-LLaVA. 
% Therefore, we suggest that the observed performance gap may stem primarily from hardware differences rather than methodological limitations.
The above results demonstrate that KTV outperforms other training-free methods with a smaller LLM on the STAR benchmark and even surpasses some training-based methods on NExTQA, EgoSchema and VideoMME, which well validates the effectiveness and efficiency of KTV for training-free video understanding.

\begin{figure*}[t] 
    \centering    
    \includegraphics[width=0.9\textwidth,height=6.25cm]{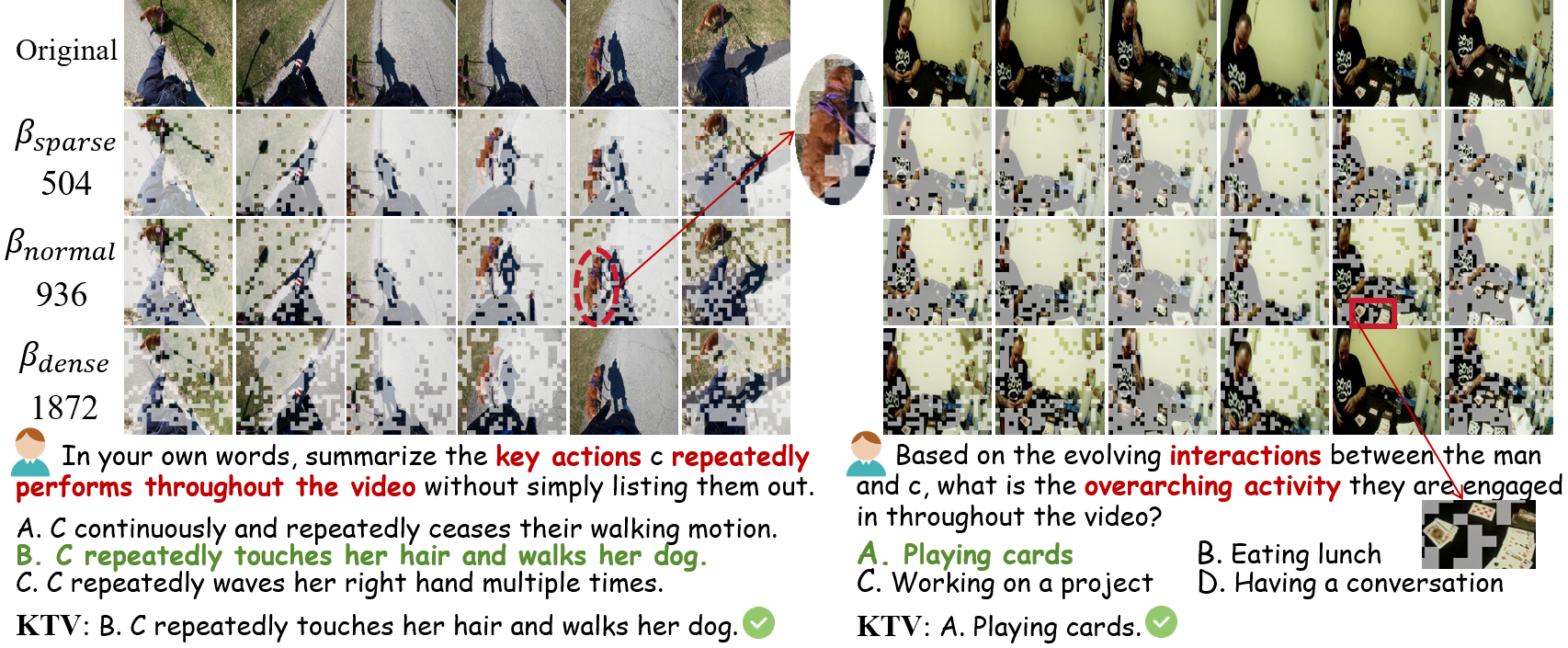}
    % \caption{\small{Visualization of KTV using different pruning rate $\beta$. The translucent patches are the pruned visual tokens.} }
    \caption{Visualization of KTV using different pruning rate $\beta$. The translucent patches are the pruned visual tokens.} 
    \label{fig: visualize}

\end{figure*}

\subsubsection{MLVU-Test}
We report the mean average accuracy (M-Avg) and subcategory-wise accuracy on MLVU-Test in Tab.\ref{tab:mlvu}.
KTV-7B-dense achieves an M-Avg of 36.9, exceeding IG-VLM-7B and SF-LLaVA-7B by $+4.8\%$ and $+4.2\%$, respectively.
KTV-34B-dense further raises M-Avg to 45.0, which not only surpasses all training-free baselines but also outperforms several training-based methods with comparable or even larger LLMs. 
For instance, it exceeds VILA-1.5-40B by $+0.8\%$, and is only marginally lower than Video-LLaMA and InternVL2 by $-0.6\%$ and $-0.7\%$, despite their use of twice the LLM size.
% Specifically, in all subcategories, KTV not only outperforms all other training-free methods but also surpasses training-based methods in some subcategories. 
In addition, across all subcategories, KTV not only consistently outperforms other training-free approaches but also surpasses many training-based methods in specific tasks.
% For example, KTV-34B-sparse achieves the highest accuracy, 53.3, of all methods in Needle QA, while KTV-34B-dense achieves the highest accuracy, 58.3, among all methods in Sports QA. 
For example, KTV-34B-sparse achieves the highest M-Avg of 53.3 in Needle QA among all methods.
KTV-34B-dense ranks first in Sports QA with 58.3 M-Avg.
Besides, KTV-7B-dense achieves the highest M-Avg of 23.3, among all methods with the smallest 7B LLM.

% This comprehensive strength underscores that our keyframe and key token selection strategy effectively preserves a rich spectrum of visual information, from overarching themes to critical minutiae, enabling robust reasoning across diverse video understanding tasks.

This comprehensive strength underscores that our keyframe and key token selection strategy effectively preserves both global and fine-grained visual cues, supporting strong reasoning across diverse video understanding tasks.

\subsubsection{Ablation Study} 

\begin{table}[!t]
    \small
    \centering
    \begin{tabular}{lccc}
    \toprule
   \textbf{Components}&\textbf{NExTQA }&\textbf{EgoSchema}&\textbf{IntentQA} \\
    
     \midrule
    \textbf{CK+IS+RS}&\textbf{\underline{65.1}} &\textbf{\underline{50.4}}&\textbf{\underline{61.3}}\\
    \midrule
    \textbf{QK+IS+RS}&63.5&46.1&60.3\\
    \textbf{CK+IS}&64.8 &49.4&61.0\\
    \textbf{CK+RS}&64.6 &48.6&60.9\\
    \textbf{RS+IS}&64.1 &48.0&60.7\\
    
    \bottomrule
    \end{tabular}

    % \caption{\small{
    % Ablation of KTV-7B-normal on NExTQA, Egoschema, and IntentQA. CK: Cluster Keyframes; QK: Question-relevant Keyframes ;IS: Importance Score; RS: Redundancy Score; 
    % }}

    \caption{{
    Ablation of KTV-7B-normal on NExTQA, Egoschema, and IntentQA. CK: Cluster Keyframes; QK: Question-relevant Keyframes; IS: Importance Score; RS: Redundancy Score; 
    }}
    \label{tab: ablation}

\end{table}

\begin{figure}[!t]
\centering
\includegraphics[width=0.82\linewidth,height=6.8cm]{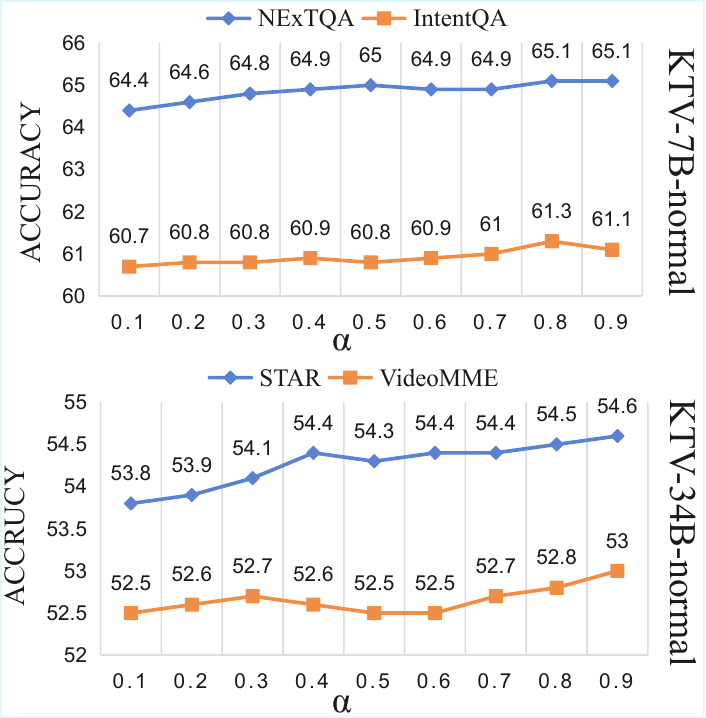}

% \caption{\small{Accuracy under different setting of $\alpha$.}}

\caption{{Accuracy under different setting of $\alpha$.}}

\label{fig: rate}
\end{figure}

% Next, we ablate the key settings of KTV. 
We conduct an ablation study to assess the key components of KTV in Tab.~\ref{tab: ablation}.
% As shown in  Tab~\ref{tab: ablation}, when we prune uninformative visual tokens relying only on importance or redundancy, the performance of both will drop. 
When visual tokens are pruned based solely on either importance or redundancy, performance consistently degrades.
For instance, on EgoSchema, removing RS and IS leads to drops of $-1.0\%$ and $-1.8\%$, respectively. 
This demonstrates the complementarity and effectiveness of combining the two pruning strategies.
Moreover, we observe that CK+IS consistently outperforms CK+RS across all benchmarks, which
% This indicates that, for VideoQA tasks, retaining tokens with high attention to the [CLS] token is slightly more beneficial than retaining globally diverse tokens.
suggests that semantic relevance plays a more crucial role than visual diversity in guiding effective token selection for VideoQA.
% Besides, only using importance score to prune visual tokens is slightly better than only using redundancy score, which illustrates that the important visual tokens that are more similar to the [CLS] token are more crucial than the non-redundant visual tokens that carry more global information in the VideoQA task.

In addition to visual token pruning, we also evaluate the efficacy of Cluster Keyframes(CK) selection. Compared with selecting keyframes based on temporal order or question, which respectively correspond to IS+RS setting and QK+IS+RS setting, our method of clustering frame-level visual features yields notable improvements, especially on long videos, \emph{e.g.}, on EgoSchema, CK achieves a significant improvement of $+2.6\%$ and $+4.3\%$ respectively. 
This gap reveals the limitation of uniform sampling, which may either missing brief yet crucial moments or redundantly sample static scenes, and question-relevant selection, which selects keyframes excessively related to the question, losing the global information.
In contrast, our clustering strategy yields visually diverse and representative frames, enabling a more comprehensive understanding of video content.

As for $\alpha$, we report accuracy under different values in Fig.\ref{fig: rate}. 
Accuracy generally increases with higher $\alpha$, peaking at values of 0.8 or 0.9. 
In addition, our method consistently outperforms the baseline and single-criterion pruning (\emph{i.e.}, importance-only or redundancy-only), highlighting the complementary strengths of the two scores.

\subsubsection{Visualization of KTV}
Fig.~\ref{fig: visualize} shows the visualization of KTV under different pruning rates $\beta$. 
Despite significantly reducing the number of visual tokens, our key token selection strategy effectively retains the most salient visual cues, ensuring that the LLM receives sufficient contextual information to produce accurate answers.  

\section{Conclusion}
In this paper, we propose \textbf{KTV}, a novel training-free, two-stage framework that addresses the critical challenges of temporal and spatial redundancy in video understanding for LLMs. 
KTV first clusters frame features to select a concise set of keyframes, then selects key visual tokens using a balanced combination of their importance and redundancy scores, which yields a concise yet highly informative representation of video content.
% Extensive experiments with LLaVA-v1.6-(7B and 34B) compared with other training-free methods and some training-based methods demonstrate that KTV achieves best or competitive performance on diverse VideoQA benchmarks while significantly reducing visual token counts and inference time. 
Extensive experiments show that KTV consistently achieves state-of-the-art or competitive performance across diverse VideoQA benchmarks, while significantly reducing the number of visual tokens and VLM inference time.
% Therefore, KTV provides valuable insights for training-free and even training-based methods in video understanding.
These results highlight KTV's potential not only for training-free approaches but also as a strong foundation for training-based video understanding methods.

% \begin{figure*}[b]  
%     \centering     
%     \includegraphics[width=0.95\textwidth]{token_prune_cropped_v2.pdf}
%     \caption{visualize} % 在 figure* 环境中直接使用 \caption
%     \label{fig:visualize}
% \end{figure*}
\section{Acknowledgments}
This work was supported by the Guangdong Province's Key Research and Development Project (2024B0101010003), the National Natural Science Foundation of China (624B2119, 62402015), the Postdocotoral Fellowship Program of CPSF (GZB20230024), the China Postdoctoral Science Foundation (2024M750100), and the Start-up Grant (No.9382010) of the City University of Hong Kong.

\bibliography{aaai2026}

\end{document}